\documentclass[letterpaper, 10 pt, conference]{ieeeconf}  

\IEEEoverridecommandlockouts                              

\usepackage{xcolor}
\usepackage{soul}

\definecolor{lightyellow}{rgb}{1, 1, 0.6}
\sethlcolor{lightyellow}
\usepackage{graphics} 
\usepackage{epsfig} 
\usepackage{times} 
\usepackage{amsmath} 
\usepackage{amssymb}  
\usepackage{booktabs} 
\usepackage{multirow} 
\usepackage[table]{xcolor}   
\usepackage{tabularx} 
\usepackage{cite} 
\usepackage{pifont}
\newcommand{\cmark}{\ding{51}} 

\usepackage{siunitx}    
\title{\LARGE \bf
ViSA-Enhanced Aerial VLN: A Visual-Spatial Reasoning Enhanced Framework for Aerial Vision-Language Navigation
}

\author{
    Haoyu Tong$^{1,2\dagger}$, 
    Xiangyu Dong$^{3\dagger}$, 
    Xiaoguang Ma$^{3,6*}$, 
    Haoran Zhao$^{4,5}$, 
    Yaoming Zhou$^{4*}$, and 
    Chenghao Lin$^{1,2}$
\thanks{The authors gratefully acknowledge the support of the Reward Funds for Research Project of Tianmushan Laboratory (No. TK-2024-D-009), the National Natural Science Foundation of China (No. 52272382), and the Key R\&D Program of Zhejiang (No. 2025C01056). This work was supported in part by the Hunan Key Scientific and Technological Research Foundation under Grant 2024QK2001, and in part by the Initiation Funding of Foshan Graduate School of Innovation, Northeastern University under Grant 200076421002.}
\thanks{$^{\dagger}$These authors contributed equally to this work.}%
\thanks{$^{*}$Corresponding authors.}%
\thanks{$^{1}$H. Tong and C. Lin are with Tianmushan Laboratory, Beihang University, Hangzhou 311115, China (e-mail: {\tt\small haoyutong@buaa.edu.cn}; {\tt\small linchenghao@buaa.edu.cn}).}%
\thanks{$^{2}$H. Tong and C. Lin are also with Hangzhou International Innovation Institute, Beihang University, Hangzhou 311115, China.}%
\thanks{$^{3}$X. Dong and X. Ma are with Foshan Graduate School of Innovation, Northeastern University, Foshan, China (e-mail: {\tt\small dxy1999ai@163.com}; {\tt\small maxg@mail.neu.edu.cn}).}%
\thanks{$^{4}$H. Zhao and Y. Zhou are with School of Aeronautic Science and Engineering, Beihang University, Beijing 100191, China (e-mail: {\tt\small zhaohaoran@buaa.edu.cn}; {\tt\small zhouyaoming@buaa.edu.cn}).}%
\thanks{$^{5}$H. Zhao is also with qingniaoAI.}%
\thanks{$^{6}$X. Ma is also with Faculty of Robot Science and Engineering, Northeastern University, Shenyang 110819, China.}%
}
\begin{document}

\maketitle
\thispagestyle{empty}
\pagestyle{empty}

\begin{abstract}
Existing aerial Vision-Language Navigation (VLN) methods predominantly adopt a detection-and-planning pipeline, which converts open-vocabulary detections into discrete textual scene graphs. These approaches are plagued by inadequate spatial reasoning capabilities and inherent linguistic ambiguities. To address these bottlenecks, we propose a Visual-Spatial Reasoning (ViSA) enhanced framework for aerial VLN. Specifically, a triple-phase collaborative architecture is designed to leverage structured visual prompting, enabling Vision-Language Models (VLMs) to perform direct reasoning on image planes without the need for additional training or complex intermediate representations. Comprehensive evaluations on the CityNav benchmark demonstrate that the ViSA-enhanced VLN achieves a 70.3\% improvement in success rate compared to the fully trained state-of-the-art (SOTA) method, elucidating its great potential as a backbone for aerial VLN systems.
\end{abstract}

\section{INTRODUCTION}

Aerial VLN requires Unmanned Aerial Vehicle (UAV) to navigate complex environments by following natural language instructions. Unlike ground-based robots confined to bounded 2D planes, UAVs operate in 3D space, where rich geometric and semantic contexts pose unique challenges in spatial reasoning and environmental
understanding \cite{Lee2025c, Liu2023c, Wang2024g, dong2025sevlnselfevolvingvisionlanguagenavigation, Anderson2018a}. 

The core difficulty of Aerial VLN lies in precisely identifying target objects and verifying their spatial relationships from a 3D perspective. Existing methods typically adopt a disjoint detection-and-planning pipeline, combining open-vocabulary detectors like Grounding DINO \cite{Liu2024k} for object recognition with discrete textual scene graphs for spatial reasoning \cite{Xu2025c, Gao2025i, Zhang2025l, Li2025k}. However, this paradigm suffers from three critical limitations.

Firstly, open-vocabulary object detectors face domain shift challenges when processing aerial view data. In unstructured urban environments, these models struggle to overcome feature mismatches caused by perspective differences, limiting their zero-shot semantic grounding capability \cite{Zhou2025d}.

Secondly, existing VLN methods rely on textual symbolic representations, such as scene graphs, for spatial relationship reasoning \cite{Xu2025c, Gao2025i}. These discrete textual representations fail to reconstruct continuous spatial layouts, rendering agents prone to relationship hallucinations, i.e., generating spatial relationship descriptions inconsistent with visual facts \cite{Wu2024e}.

Thirdly, spatial descriptions in natural language possess inherent semantic ambiguity \cite{Hong2024}. The accurate parsing of prepositions, e.g., ``between'' or ``across from'', depends heavily on the reference frame of the visual context, while discrete text modalities fail to capture such continuous spatial constraints \cite{Liu2023d} and cannot utilize this visual continuity to disambiguate instruction understanding.

While VLMs can directly perform visual reasoning on aerial images, recent research illustrates their fundamental defects in spatial cognition \cite{Chen2026}. Unique top-down perspectives and scale variations in Aerial VLN amplify these limitations, making VLN models trained primarily on ground data struggle with aerial perception. Specifically, VLMs exhibit limited spatial acuity during complex geographic queries \cite{Alam2026} and reasoning inconsistencies under cross-modal constraints \cite{Zhu2026}. Additionally, VLMs in complex urban scenes produce object relationship hallucinations \cite{Wu2024e} and struggle to distinguish between visually similar candidates \cite{Zhang2024d}.

While visual prompting techniques succeed in general 2D image understanding, their applications in dynamic aerial navigation remains an open challenge, and a collaborative architecture to bridge the gap between static object tagging and active spatial reasoning in 3D spatial scenes is highly desirable. Inspired by the proven efficacy of Visual Referencing Paradigms, particularly the explicit region-based prompting principles popularized by Set-of-Mark (SoM)\cite{Yang2023c}, we propose a Visual-Spatial Reasoning (ViSA) enhanced framework for aerial VLN, Specifically, a Visual Prompt Generator first transforms raw aerial observations into structured and region-annotated visual representations. Building on this, a Verification Module performs explicit spatial reasoning directly on image planes to mitigate relationship hallucinations. Finally, a Semantic-Motion Decoupled Executor is designed to separate these high-level semantic decisions from precise motion control, incorporating landmark-based waypoint generation to plan efficient exploration paths.


The main contributions of this work are summarized as follows:
\begin{itemize}
    \item We propose the ViSA-enhanced Aerial VLN framework, a zero-shot architecture that mitigates spatial reasoning hallucinations in language-goal aerial navigation by restructuring the task into three distinct phases: Perception, Verification, and Execution.
    
    \item A Visual Prompt Generator is introduced to partition raw images into regions of varying granularity with SoM annotations, providing the VLM with a structured visual representation for precise spatial analysis.
    
    \item A Verification Phase is proposed to implement explicit Three-Stage Verification Reasoning. This strictly grounds spatial logic within the visual modality, ensuring superior performance over conventional text-centric spatial reasoning approaches.
    
    \item A Semantic-Motion Decoupled Executor (Executor) is designed to bridge high-level semantic decisions with low-level discrete actions via landmark-based waypoint generation and specialized task primitives.
    
    \item Comprehensive evaluations on the CityNav benchmark demonstrate that our zero-shot approach achieves a 70.3\% improvement on success rate over the fully trained SOTA on the Test-Unseen split.
\end{itemize}

\begin{figure*}[htbp]
    \centering
    \includegraphics[width=\linewidth]{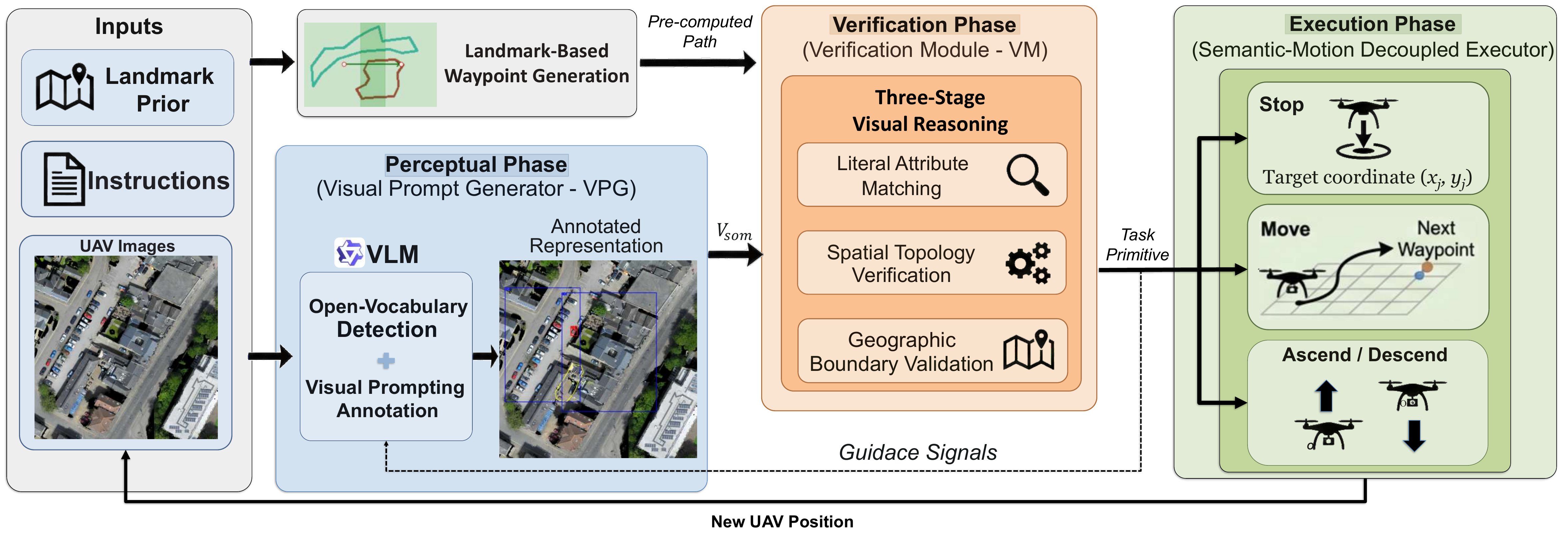}
    \caption{Overall architecture of \textbf{ViSA}. The system iterates through three phases per time step: (1) \textbf{VPG} transforms raw observations into SoM-annotated visual representations; (2) \textbf{VM} performs Three-Stage Verification reasoning and emits task primitives; (3) \textbf{Executor} translates semantic decisions into UAV motion. Closed-loop feedback from VM to VPG enables adaptive detection across iterations.}
    \label{fig:architecture}
  \end{figure*} 

\section{RELATED WORK}

\subsection{Aerial Vision-Language Navigation (VLN)}


Ground-level VLN, pioneered by the R2R benchmark \cite{Anderson2018a}, has driven innovations in cross-modal alignment, spatial reasoning, and pre-training paradigms~\cite{Zhang2024a}. In contrast, aerial VLN faces distinct challenges from aerial viewpoints and continuous 3D action spaces. Pioneering benchmarks~\cite{Liu2023c, Fan2023a, Wang2024g}  demonstrate that directly using ground-based paradigms on aerial VLN leads to poor generalization, emphasizing the need for tailored approaches. To enhance its reasoning and navigation capabilities, recent works have integrated Multimodal Large Language Models (MLLMs) into aerial VLN~\cite{Xu2025c, Cai2025b, Gao2025i, Zhang2025l, Wang2024g}. These methods either adopt a disjoint detection-and-planning pipeline that restricts holistic reasoning or require resource-intensive pre-training. In contrast, our work introduces a triple-phase collaborative architecture that leverages structured visual prompting to unlock the spatial reasoning potential of VLMs without the need for additional training or complex intermediate representations.

\subsection{Visual Prompting}

Visual Prompting (VP) has emerged as a core non-parametric interaction paradigm for enhancing the reasoning and grounding capabilities of VLMs, eliminating the need for extensive fine-tuning \cite{wu2024visual}. Recent advances illustrate that augmenting visual inputs with explicit markers, such as drawn shapes, bounding boxes, or segmentation masks, can direct model attention and facilitate user-aligned comprehension \cite{lin2024draw, Yang2023c, jiang2024joint}. Furthermore, techniques like Contrastive Region Guidance \cite{wan2024contrastive} and Visual Position Prompting \cite{tang2026visual} have proven that training-free visual manipulations and explicit coordinate encodings can improve region-specific grounding. Recently, VP has been introduced into VLN to guide agents via explicit visual cues~\cite{Cai2024,Feng2025b}. However, existing approaches primarily utilize VP for immediate action guidance, leaving the potential of VP for enhancing spatial reasoning in aerial VLN unexplored.

\section{METHODOLOGY}
\label{sec:methodology}
\subsection{Problem Formulation and Framework Overview}

We formulate aerial VLN tasks in unknown urban environments as follows. Prior information of landmarks $L \in \mathcal{L}$ is accessible to the UAV, with prior knowledge defined as a tuple $\mathcal{K}_L = (\mathcal{I}_{label}, \mathcal{P}, \mathcal{C})$, where $\mathcal{I}_{label}$ denotes the landmark name, $\mathcal{P}$ denotes the centroid coordinates, and $\mathcal{C}$ denotes the contour point set. Given an instruction $\mathcal{T}$ (e.g., ``house with white roof on the left of Broadway road''), the UAV obtains the location and shape information of $L$ through $\mathcal{I}_{label}$.

Formally, each navigation episode is defined as a triplet $\mathcal{E} = (\mathcal{T}, p_0, \mathcal{K}_{\text{prior}})$, where $\mathcal{T}$ denotes the natural language instruction, $p_0 = (x_0, y_0, z_0, \theta_0)$ denotes the initial pose of the UAV (with $(x_0, y_0)$ as the world coordinates, $z_0$ as the altitude, and $\theta_0$ as the heading), and $\mathcal{K}_{\text{prior}} = \{\mathcal{K}_{L^{(1)}}, \dots, \mathcal{K}_{L^{(N)}}\}$ denotes the set of prior knowledge for all available landmarks.

At each time step $t$, the UAV obtains a bird's-eye view observation image $I_t$ and the current pose $p_t = (x_t, y_t, z_t, \theta_t)$. The CityNav environment natively provides a low-level discrete action space $\mathcal{A}_{\text{low}}$ including move-forward, turn-left, turn-right, ascend, descend, stop, and our framework employs a hierarchical control strategy to bridge the gap between abstract visual reasoning and precise flight. In fact, our system issues a high-level semantic task primitive $a_t \in \{\text{stop}, \text{move}, \text{ascend}, \text{descend}\}$, which will be translated into a sequence of low-level $\mathcal{A}_{\text{low}}$ commands by the Executor subsequently, as detailed in Section~\ref{sec:Executor}.

The objective is to reach the target location $p_{target}$ within a limited number of steps $T$. Success is determined by the Euclidean distance $\|p_T^{\text{pos}} - p_{\text{target}}\| \leq \epsilon$ between the termination position and the target, where $\epsilon = 20$ meters is the success threshold and $p_T^{\text{pos}}$ denotes the world coordinates of the termination position.

Given an instruction $\mathcal{T}$ and landmark priors $\mathcal{K}_{\text{prior}}$, the framework invokes landmark-based waypoint generation to pre-compute an exploration path from merged landmark contours as depicted in Figure~\ref{fig:architecture}. At each time step $t$, the UAV acquires a raw bird's-eye view image $I_t$ and proceeds through three distinct, tightly coupled phases as shown in Sections~\ref{sec:Perception} to \ref{sec:Executor}. By restructuring the navigation task into these three phases, ViSA mitigates VLM hallucinations and keeps spatial reasoning strictly within the visual modality. A closed-loop feedback mechanism connects the phases, i.e., when the Verification Phase yields insufficient evidence, a natural-language guidance signal $g$ feeds back to steer the next round of perception.

\subsection{Landmark-Based Waypoint Generation}
For instructions involving complex spatial constraints (e.g., ``around the block''), the Executor leverages prior landmark contour geometry to pre-compute efficient exploration waypoints.

Given prior contours of instruction-related landmarks, the waypoint generation proceeds in three steps: 

~\textbf{Contour merging}: compute the merged contour as
\begin{equation}
\mathcal{C}_{\text{merged}} = \bigcup_{i=1}^N \mathcal{C}^{(i)}
\end{equation}
and identify intersection areas as priority search regions; 

~\textbf{Grid generation}: produce candidate observation points within the merged contour's bounding box with step size
\begin{equation}
\Delta_{\text{grid}} = \text{FOV}(z) \times (1 - r)
\end{equation}
where the field of view (FOV) at altitude $z$ is given by
\begin{equation}
\text{FOV}(z) = (z - z_{\text{ground}}) \times \alpha
\end{equation}
with parameters $\alpha = 2.0$ and $r = 0.3$; 

~\textbf{Greedy selection}: apply a greedy set cover to select an observation subset $\mathcal{P}_{\text{selected}}$ that maximizes cumulative coverage over $\mathcal{C}_{\text{merged}}$ with FOV overlap constraints. Finally, visitation order is optimized via Traveling Salesperson Problem (TSP) to minimize total flight distance. The Executor generates this path during initialization and executes \textbf{move} commands sequentially along pre-computed waypoints.

\subsection{Perception Phase}\label{sec:Perception}
Within the Perception Phase, we employ a Visual Prompt Generator (VPG) to process the raw bird's-eye view image $I_t$. To extract semantic regions, the VPG leverages the native open-vocabulary detection capabilities of modern VLMs. Unlike traditional open-vocabulary detection models that are implicitly bounded by their training distributions, restricted to isolated word queries, and lack holistic environmental sensitivity, VLM-based detection offers superior adaptability. It accommodates complex, free-form natural language prompts to simultaneously localize multiple diverse objects in a single pass with contextual awareness. Upon identifying these candidate targets, the VPG partitions the image into regions of varying granularity and overlays them with SoM annotations, producing a structured visual representation $\mathcal{V}_{\text{som}} \triangleq (I_{\text{som}},\, M_{\text{som}})$. Here, $I_{\text{som}}$ is the annotated image and $M_{\text{som}}$ is the symbolic mapping:
\begin{equation}
M_{\text{som}} = \{[1] \rightarrow o_1,\; [2] \rightarrow o_2,\; \ldots,\; [N] \rightarrow o_N\},
\end{equation}
where each numerical ID uniquely references a physical entity.

This process provides the VLMs with a structured visual representation for precise spatial analysis. By adopting a high-recall open-vocabulary detection strategy, the VPG captures all potential candidates in the scene. Rather than prematurely filtering entities based on heuristic confidence thresholds, the Perception Phase passes all detected proposals to the Verification Phase, thereby decoupling detection recall from verification precision.

\subsection{Verification Phase}\label{sec:Verification}
Building upon the annotated images and structured visual representations provided by the Perception Phase, the Verification Module (VM) executes explicit Three-Stage Verification Reasoning. The VM receives $\mathcal{V}_{\text{som}}$ together with the instruction $\mathcal{T}$ and outputs a verification decision along with an optional search guidance signal $g$. By reasoning directly on image planes, this phase strictly grounds spatial logic within the visual modality, ensuring superior performance over conventional text-centric spatial reasoning. This phase decomposes spatial verification into three sequential stages:

\textbf{Literal Attribute Matching.} By referencing SoM identifiers (e.g., \ding{172}), the VM verifies whether the visible features of candidate targets align with the natural language instruction. It adheres strictly to literal observation; the VM evaluates only directly visible attributes and defers decisions by marking targets as ``pending'' when visual evidence remains insufficient.

\textbf{Spatial Topology Verification.} The VM verifies spatial relationships by referencing numerical IDs (e.g., ``\ding{172} is across from \ding{173}'') rather than relying on discrete textual scene graphs, eliminating referential ambiguity.

\textbf{Geographic Boundary Validation.} The VM ensures candidates are geographically legitimate by checking whether they conform to the spatial relationships with known landmarks, avoiding selecting targets that visually match but are geographically incorrect.

\textbf{Iterative VPG Prompting.} When the current scene yields inconclusive spatial evidence, the VM outputs a natural-language guidance signal $g$ (e.g., ``focus on white vehicles near the intersection''). This signal is directly passed back to the Perception Phase as the textual prompt for the subsequent detection round, helping narrow the search scope.

\subsection{Execution Phase}\label{sec:Executor}

The Execution Phase serves as the interface between abstract semantic reasoning and physical control. To circumvent the limitations of zero-shot VLMs in generating low-level control signals, we introduce a Semantic-Motion Decoupled Executor. We define the execution mapping as:
\begin{equation}
f_{\text{exec}}: (I_t, p_t, a_t, \mathcal{K}_{\text{prior}}) \rightarrow p_{t+1}
\end{equation}

which translates the current state $s$, a specific task primitive $a_t$, and the available landmark priors $\mathcal{K}_{\text{prior}}$ into the target UAV pose $p_{t+1}$.

The VM's semantic decisions are mapped to UAV actions through three specialized task primitives:

\textbf{Stop.} Upon target confirmation by the VM, the Executor must translate the visual semantic result into a physical navigation goal. To map the 2D pixel centroid of the target to 3D world coordinates, we adopt the unprojection formulation from GeoNav~\cite{Xu2025c}. By leveraging the camera intrinsic matrix and the known UAV flight altitude relative to the ground plane, the pixel centroid is directly unprojected to obtain the target's world coordinates $p_{t+1} = (x_{\text{target}}, y_{\text{target}}, z_t, \theta_t)$. The Executor then navigates the UAV directly to this location, effectively avoiding multi-step accumulated errors.

\textbf{Move.} The Executor flies the UAV to the next pre-computed waypoint from $\mathcal{P}_{\text{selected}}$, where the Perception Phase is triggered to re-observe the scene and autonomously decide the next action.

\textbf{Ascend / Descend.} The Executor adjusts altitude by $\Delta z = \pm 10$ meters to expand or shrink the observation FOV accordingly, balancing detail resolution with scene coverage.

\section{Experiment}
\label{sec:experiments}

\begin{table*}[thp]
\centering
\caption{Performance comparison of zero-shot methods on the CityNav dataset across three difficulty levels. \textbf{Bold} indicates best, \underline{underline} indicates second-best.}
\label{tbl:zs_results}
\setlength{\tabcolsep}{3.5pt}
\resizebox{\textwidth}{!}{
\begin{tabular}{@{}l cccc cccc cccc@{}}
\toprule
\multirow{2}{*}{\textbf{Method}} & \multicolumn{4}{c}{\textbf{Easy}} & \multicolumn{4}{c}{\textbf{Medium}} & \multicolumn{4}{c}{\textbf{Hard}} \\
\cmidrule (lr){2-5} \cmidrule (lr){6-9} \cmidrule (lr){10-13}
 & NE$\downarrow$ & SR$\uparrow$ & OSR$\uparrow$ & SPL$\uparrow$ & NE$\downarrow$ & SR$\uparrow$ & OSR$\uparrow$ & SPL$\uparrow$ & NE$\downarrow$ & SR$\uparrow$ & OSR$\uparrow$ & SPL$\uparrow$ \\
\midrule
Random & 340.62 & 0.00 & 0.00 & 0.00 & 548.30 & 0.00 & 0.00 & 0.00 & 654.26 & 0.00 & 0.00 & 0.00 \\
Greedy & 105.63 & 1.55 & 3.15 & 0.97 & 103.00 & 0.00 & 4.88 & 0.00 & 78.80 & 0.00 & 2.71 & 0.00 \\
GPT-4o & 301.45 & 0.00 & 7.50 & 0.00 & 327.25 & 0.00 & 6.25 & 0.00 & 401.63 & 0.00 & 0.00 & 0.00 \\
Qwen3-VL-PLUS & 838.33 & 0.49 & 7.54 & 0.03 & 836.11 & 0.00 & 2.49 & 0.00 & 876.23 & 0.24 & 1.65 & 0.17 \\
\midrule
GeoNav~\cite{Xu2025c} & \underline{59.86} & \underline{26.53} & \textbf{73.47} & \underline{12.05} & \underline{53.80} & \underline{22.92} & \textbf{39.58} & \underline{17.06} & \underline{68.90} & \underline{16.67} & \underline{22.92} & \underline{12.49} \\
\rowcolor{gray!15}
\textbf{ViSA-enhanced VLN (Ours)} & \textbf{47.75} & \textbf{30.19} & \underline{38.39} & \textbf{19.35} & \textbf{51.51} & \textbf{29.34} & \underline{34.45} & \textbf{23.94} & \textbf{58.96} & \textbf{28.54} & \textbf{34.97} & \textbf{24.33} \\
\bottomrule
\end{tabular}
}
\end{table*}

\subsection{Experiment Setting}
\subsubsection{Dataset Preparation}
The navigation instructions utilized in our experiments are derived from the original and refined CityNav \cite{Lee2025c} dataset, which comprises 32,326 natural language descriptions paired with human demonstration trajectories, all collected by crowd-sourcing. Each language description encompasses information such as landmarks, regions, and objects. In each of the Validation Seen (Val-Seen), Validation Unseen (Val-Unseen), and Test-Unseen datasets, tasks are divided into three levels of difficulties with increasing start-target distances. The UAV images are taken from SensatUrban \cite{hu2022sensaturban}, which gathers orthographic projections and depth maps of 13 blocks in Birmingham and 33 blocks in Cambridge. These data are utilized to simulate the RGBD inputs that an actual UAV acquires during navigation. Additionally, the geometric outlines of landmarks within the geographic information database are obtained from CityRefer \cite{miyanishi2023cityrefer}, providing essential information for the navigation tasks.

\subsubsection{Evaluation Metrics}
Following standard evaluation protocol established by CityNav, we employ four metrics to comprehensively assess navigation performance.

\begin{itemize}
\item[$\bullet$] \textbf{Navigation Error (NE):} The Euclidean distance between the agent's final position and the ground truth target location. Lower NE indicates superior localization accuracy.

\item[$\bullet$] \textbf{Success Rate (SR):} The percentage of episodes in which the agent terminates within a predefined success threshold (20 meters) of the target, reflecting goal-reaching capability.

\item[$\bullet$] \textbf{Oracle Success Rate (OSR):} The proportion of episodes where the agent, at any point during navigation, approaches within 20 meters of the target regardless of final stopping position. This metric decouples instruction grounding from path execution.

\item[$\bullet$] \textbf{Success weighted by Path Length (SPL):} SPL is a metric where SR is weighted by the ratio of the optimal path length to the actual path taken, rewarding short navigation paths.
\end{itemize}

\subsubsection{Implementation Details}
Our method is deployed on an NVIDIA GeForce RTX 4090 GPU with 24GB memory and an Intel Xeon Silver 4210R CPU. Unlike previous pipelined approaches that rely on specialized detector models, e.g., Grounding DINO for perception, our framework leverages the native object detection capabilities of VLM. Specifically, the VPG directly utilizes the Qwen3-VL-PLUS online API to propose open-vocabulary bounding boxes. The same VLM is subsequently utilized by the Verification Module for spatial reasoning, using a sampling temperature of 0.6. To ensure a strictly fair evaluation, baseline methods GeoNav and FlightGPT are used with the exactly same landmark priors ($\mathcal{K}_{\text{prior}}$) from the CityRefer database as our proposed framework. 

\subsection{Main Results}
\subsubsection{Comparison with Zero-Shot Methods}
Table~\ref{tbl:zs_results} presents the overall performance of our method compared to \textbf{zero-shot} baseline methods on the CityNav Val-Seen split. The ViSA method achieves the best SR, NE, and SPL metrics across all difficulty levels.

Specifically, ViSA obtains SRs of 30.19\%, 29.34\%, and 28.54\%, surpassing the primary baseline GeoNav by 13.8\%, 28.0\%, and 71.2\% for Easy, Medium, and Hard tasks, respectively. This progressive improvement demonstrates that our visual-centric reasoning paradigm maintains robust target grounding without relying on heavy and error-prone explicit scene-graph generation as spatial complexity increasing.

Furthermore, our method achieves superior path efficiency, evidenced by consistent improvements in SPL across all difficulty levels. While, the GeoNav exhibits a relatively high OSR (e.g., 73.47\% on Easy level), it suffers from a substantial discrepancy between its OSR and actual SR (26.53\%). This indicates that although the GeoNav frequently encounters the target during its navigation trajectory, it lacks the recognition capacity to definitively halt and confirm it. In contrast, ViSA maintains a narrow margin between OSR and SR (e.g., 38.39\% OSR vs. 30.19\% SR on Easy level). This minimal discrepancy validates that ViSA enhances the agent's target grounding and confirmation capabilities. In fact, once the destination falls within the FOV, it is reliably identified and explicitly confirmed.

To evaluate the stability of ViSA against the inherent stochasticity of VLMs, we conduct five independent runs across all difficulty levels. The SR exhibits a standard deviation of merely $0.6\%$ on Easy tasks and $1.3\%$ on Hard tasks. Furthermore, both NE and SPL maintain tight bounds across all splits ($\le 1.0$m and $\le 1.1\%$, respectively). This minimal variance confirms that our explicit visual grounding and Three-Stage Verification Reasoning effectively constrain the VLM's reasoning process, largely mitigating the "lucky seed" phenomenon and ensuring highly reproducible navigation performance.


\begin{table}[thp]
\caption{Performance comparison of trained methods on the CityNav \textbf{Test-Unseen} split. \textbf{Bold} indicates best, \underline{underline} indicates second-best.}
\label{tbl:trained_results}
\setlength{\tabcolsep}{4pt}
\begin{tabular*}{\linewidth}{@{\extracolsep{\fill}}l cccc@{}}
\toprule
\textbf{Method} & NE$\downarrow$ & SR$\uparrow$ & OSR$\uparrow$ & SPL$\uparrow$ \\
\midrule
Random & 208.80 & 0.00 & 1.44 & 0.00 \\
Seq2Seq & 174.50 & 1.73 & 8.57 & 1.69 \\
CMA & 179.10 & 1.61 & 10.07 & 1.57 \\
MGP & 93.80 & 6.38 & 26.04 & 6.08 \\
FlightGPT~\cite{Cai2025b} & \underline{76.20} & \underline{21.20} & \underline{35.38} & \underline{19.24} \\
\midrule
\rowcolor{gray!15}
\textbf{ViSA-enhanced VLN (Ours)} & \textbf{45.73} & \textbf{36.11} & \textbf{43.37} & \textbf{27.31} \\
\bottomrule
\end{tabular*}
\end{table}

\subsubsection{Comparison with Supervised Methods}
Table~\ref{tbl:trained_results} presents the performance comparison between our zero-shot method and supervised learning-based approaches on the CityNav Test-Unseen split. ViSA achieves superior performance against all trained baselines without requiring task-specific fine-tuning.

ViSA attains a SR of 36.11\% and an SPL of 27.31\%, outperforming traditional learning-based methods such as Seq2Seq (SR: 1.73\%, SPL: 1.69\%) and CMA (SR: 1.61\%, SPL: 1.57\%). Compared to the MGP baseline, which incorporates map-based guidance, our method improves SR by 466\% and SPL by 349\%. This margin demonstrates the superiority of our vision-centric reasoning paradigm over traditional map-dependent representations in handling unseen environments.

Notably, ViSA surpasses FlightGPT~\cite{Cai2025b}, the previous SOTA supervised method that employs extensive Supervised Fine-Tuning (SFT) and Reinforcement Learning (RL) training by 70.3\% and 41.9\% for SR and SPL, respectively. This result demonstrates that appropriate architectural design, i.e., structured visual prompting and explicit spatial verification, enables general-purpose VLMs to outperform specialized models trained on domain-specific data. Furthermore, the lower NE of our method (45.73m compared to FlightGPT's 76.20m) validates the precision of our landmark-based waypoint generation and closed-loop verification mechanism.

By decoupling perception and verification through our dual-module architecture, and by leveraging visual continuity via visual prompting, our method achieves superior generalization without the distribution shift risks inherent to supervised training. In fact, higher OSR (43.37\% vs. 35.38\%) indicates that our targeted exploration strategy maintains broader coverage while simultaneously improving localization accuracy, seamlessly translating into the 14.91\% absolute gain in final SR.

\begin{figure*}[htbp]
    \centering
    \includegraphics[width=\textwidth]{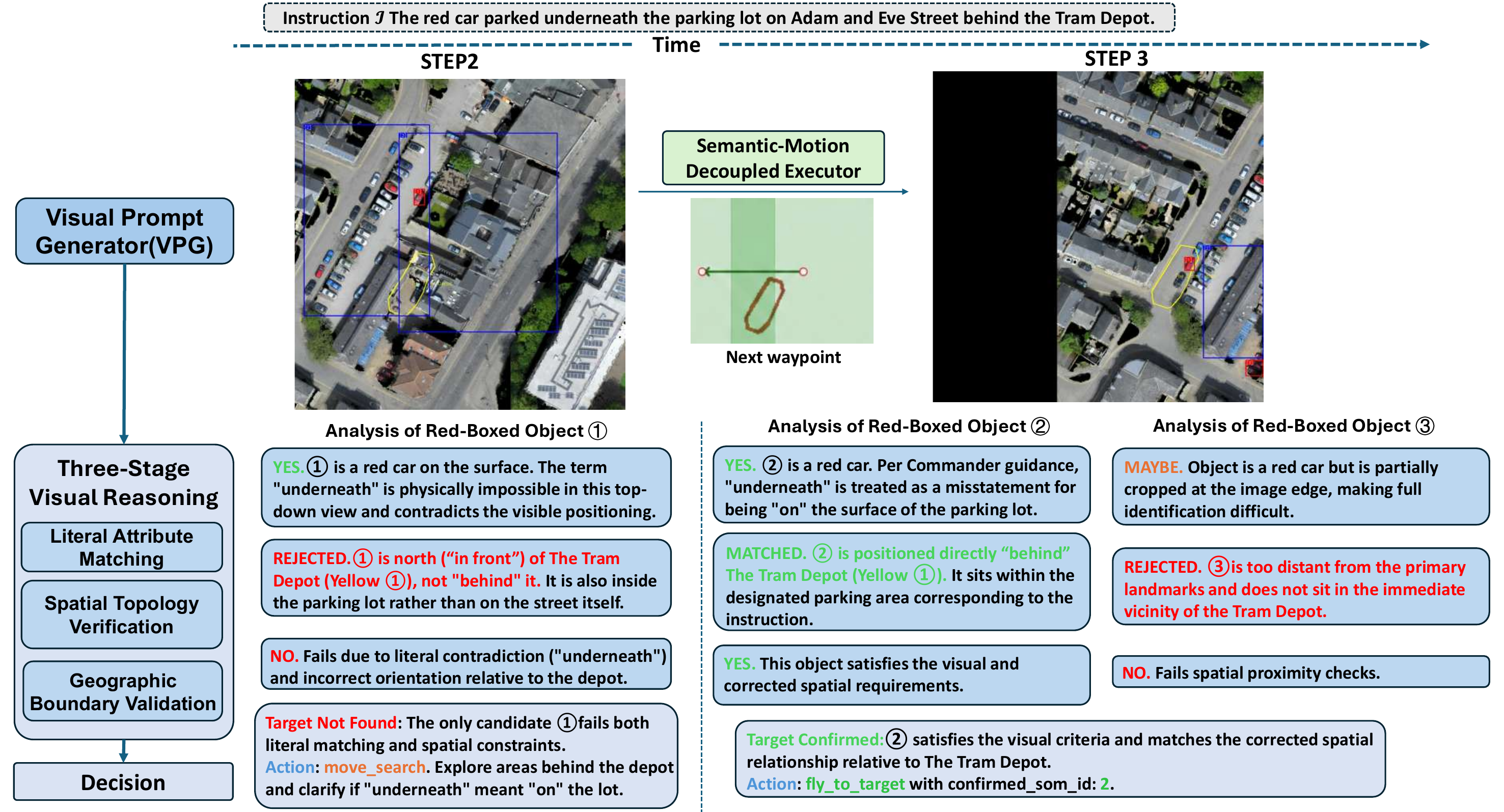}
    \caption{Visualization of the VLM's explicit reasoning trace during a navigation episode in the CityNav environment.ViSA demonstrates robust spatial reasoning by accommodating a flawed spatial preposition (``underneath'' instead of ``on''). The sequence illustrates the logical rejection of an initial false-positive red car \ding{172} located in \textit{front} of the Tram Depot \ding{174}, followed by closed-loop guided search that successfully locates and verifies the correct target positioned \textit{behind} the depot.}
    \label{fig:qualitative}
\end{figure*}

\subsection{Ablation Study}
To systematically validate the effectiveness of our triple-phase architecture, we conduct ablation tests on the Val-Seen Easy.

\textbf{Role of the Visual Prompting and Verification Reasoning.}
To evaluate the impact of structured VP, we introduce a \textit{w/o V} variant that directly feeds raw, unannotated bird's-eye view images to the VLM, bypassing the VP step. As shown in Table~\ref{tbl:ablation}, removing the VPG degrades navigation capability, with the SR dropping from 30.19\% to 20.83\%. The absence of VP exposes the VLM to severe referential ambiguity. Even when equipped with spatial reasoning, the model struggles to ground complex spatial constraints to specific pixel regions in dense urban scenes. This confirms that VP is a fundamental prerequisite for accurate spatial analysis. 

Furthermore, we construct a \textit{w/o R} variant by replacing our Three-Stage Verification Reasoning with unconstrained, direct VLM generation, allowing the model to select targets without sequential logical checks. This modification reduces the SR to 20.14\% and the OSR to 29.03\%. We observe that without the structured reasoning pipeline, the rationales provided by the VLM for its navigation decisions are often insufficient and logically incomplete. The model frequently fails to account for the exhaustive set of spatial constraints mentioned in the instructions, often prioritizing a single salient attribute while overlooking the complex ensemble of topological and geographic requirements. Our explicit reasoning pipeline forces the model to anchor its logic strictly on visual facts, effectively suppressing these hallucinations. 

Notably, simultaneously removing both components (\textit{w/o R+V}) results in a performance collapse (SR: 10.83\%, OSR: 24.17\%). This sharp decline demonstrates that VP and explicit spatial reasoning are highly complementary and mutually dependent. Visual prompting provides the necessary addressable entities, while structured reasoning utilizes these entities to enforce logical spatial constraints.

\textbf{Role of the Dual-Stage VLM Inference.} 
Our triple-phase framework employs a decoupled VLM design that separates visual perception and spatial verification into two distinct inference stages. To justify our decoupled design, we test a \textit{w/o D} variant where a single VLM is prompted to simultaneously perform open-vocabulary detection and spatial verification in one pass. This results in a significant performance collapse, with SR decreasing to 20.56\% and OSR decreasing to 22.50\%. Forcing a single VLM to handle both dense visual perception (identifying dozens of objects) and complex multi-step logical reasoning overwhelms the model's context window and cognitive load. The model's attention becomes diluted, often causing it to hallucinate nonexistent objects or prematurely terminate the search. By decoupling perception from verification into two phases, ViSA successfully distributes this cognitive burden.

\textbf{Role of the Execution Phase.}
Finally, we evaluate the \textit{w/o E} variant by removing the Executor, forcing the VLM to directly output low-level discrete actions $\mathcal{A}_{\text{low}}$ (e.g., turn-left, move-forward). Under this setting, the navigation almost entirely fails. Current VLMs inherently lack the capacity to map visual semantics into low-level discrete actions and infer complex spatiotemporal relationships. In fact, the VLM frequently falling into invalid, repetitive action loops (e.g., issuing continuous turn-left commands) and failing to formulate any systematic exploration strategy. The Executor is indispensable because it successfully bridges the fundamental gap between the VLM's high-level semantic reasoning and the UAV's low-level flight control.

\begin{table}[!thp]
\centering
\small 
\caption{Ablation study on Val-Seen Easy. \cmark{} indicates the component is included. Abbreviations: \textbf{V}: Visual Prompting, \textbf{R}: Verification Reasoning, \textbf{D}: Dual-stage VLM, \textbf{E}: Executor.}
\label{tbl:ablation}
\begin{tabularx}{\linewidth}{@{} 
  l 
  @{\hspace{12pt}} c @{\hspace{6pt}} c @{\hspace{6pt}} c @{\hspace{6pt}} c 
  @{\hspace{15pt}} *{4}{>{\centering\arraybackslash}X} 
@{}}
\toprule
\textbf{Method} & \textbf{V} & \textbf{R} & \textbf{D} & \textbf{E} &
\textbf{NE}$\downarrow$ & \textbf{SR}$\uparrow$ & \textbf{OSR}$\uparrow$ & \textbf{SPL}$\uparrow$ \\
\midrule
ViSA & \cmark & \cmark & \cmark & \cmark & \textbf{47.75} & \textbf{30.19} & \textbf{38.39} & \textbf{19.35} \\
w/o R & \cmark & & \cmark & \cmark & 77.68 & 20.14 & 29.03 & 13.02 \\
w/o V & & \cmark & \cmark & \cmark & 54.30 & 20.83 & 33.19 & 10.44 \\
w/o R+V & & & \cmark & \cmark & 63.24 & 10.83 & 24.17 & 5.90 \\
w/o D & \cmark & \cmark & & \cmark & 57.48 & 20.56 & 22.50 & 17.74 \\
w/o E & \cmark & \cmark & \cmark & & 88.90 & 9.51 & 10.73 & 8.99 \\
\bottomrule
\end{tabularx}
\end{table}

\subsection{Qualitative Analysis}
\label{sec:qualitative}

To demonstrate the efficacy of ViSA in handling complex spatial instructions and textual ambiguities, we present a successful high-precision navigation episode in the Cambridge environment with terminal Navigation Error of 0.18m. 

\textbf{Instruction Analysis and Waypoint Generation.} The task requires the UAV to locate: a red car parked underneath the parking lot on Adam and Eve Street behind the Tram Depot.(Note: The instruction contains a flawed preposition, stating ``underneath'' instead of ``on'' the parking lot). The Executor extracts the geographic contour of ``The Tram Depot'' from the prior database and computes a focused exploration path, preventing exhaustive global search.

\textbf{Initial Perception and Logical Rejection.} At the initial observation point, the VPG leverages its open-vocabulary capacity to identify and apply SoM identifiers to a candidate red car \ding{172} and the Tram Depot building \ding{174}. The VM then receives the structured visual representation and performs Three-Stage Verification Reasoning:
\begin{enumerate}
    \item \textit{Literal Attribute Matching:} The VM successfully grounds the visual attributes ``red'' and ``car'' to candidate \ding{172}. Furthermore, it detects the physical impossibility of ``underneath'' the parking lot, accommodating this flawed preposition by implicitly grounding the search to the visible surface based on visual common sense.
    \item \textit{Spatial Topology Verification:} The VM explicitly evaluates the local topology and determines that candidate \ding{172} is positioned in \textit{front} of the Tram Depot \ding{174}, directly contradicting the ``behind'' requirement.
    \item \textit{Geographic Boundary Validation:} Beyond local topology, the VM verifies the candidate against macro-level geographic constraints. It recognizes that candidate \ding{172} resides on the main thoroughfare rather than within the specific boundaries of the ``parking lot on Adam and Eve Street'', violating the geographic requirement.
\end{enumerate}
Consequently, the VM rejects this false-positive candidate, sends a closed-loop guidance signal back to the VPG (``Search for a red car positioned behind \ding{174}, not in front''), and triggers a \textbf{move} primitive to continue exploration.

\textbf{Guided Search and Target Confirmation.} When reaching the next waypoint, the VPG adjusts its attention based on the feedback and successfully locks onto a new red car \ding{172} behind the depot. The VM performs another rigorous three-stage verification, i.e., the candidate visually matches a red car, correctly sits directly behind the Tram Depot, and falls within the geographic boundaries of the  parking lot on Adam and Eve Street. With all constraints satisfied, the VM emits the stop decision, and the Executor navigates the UAV precisely to the target coordinates.

\section{CONCLUSION AND FUTURE WORK}
\label{sec:conclusion}

In this paper, we present ViSA-enhanced VLN, a zero-shot VLN framework that redefines spatial reasoning for Aerial VLN. ViSA circumvents the information bottlenecks prevalent in discrete textual scene graphs by confining spatial reasoning strictly within the visual modality via visual prompting and a three-phase architecture. The Perception Phase and Verification Phases synergize high-recall visual perception with rigorous explicit reasoning, while the Execution Phase translates these semantic decisions into efficient flight paths via a Semantic-Motion Decoupled Executor. Experiments on the CityNav benchmark demonstrate that ViSA outperforms existing zero-shot and fully-trained VLN baselines.

Several limitations persist and warrant future investigation. First, the reliance on large VLM APIs introduces inference latency, restricting real-time deployment on low-power edge devices. Second, although our framework utilizes altitude adjustments, it lacks an active 3D perception mechanism such as lateral maneuvers or camera pitch control to fully resolve the occlusions of vertical features like building facades. Third, our current waypoint generation relies on landmark priors ($\mathcal{K}_{\text{prior}}$) as dictated by the CityNav benchmark which degrades transferability. To address these limitations, future work will focus on three directions: (1) deploying lightweight VLMs via model compression to reduce edge inference latency; (2) incorporating active 6-DoF camera control strategies to overcome 3D visual occlusions; and (3) integrating multimodal world models to eliminate reliance on landmark priors, thereby enabling fully autonomous exploration in completely unmapped environments.


\bibliography{IEEEabrv, references}

\end{document}